\documentclass{article}
\usepackage{spconf,amsmath,graphicx}
\usepackage{epsfig}
\usepackage{amssymb}
\usepackage{subfig}
\usepackage{cite}
\usepackage{multirow}
\usepackage{epstopdf}
\usepackage{graphicx}
\usepackage{amsmath}
\usepackage{color, soul}
\usepackage[style=base]{caption}
\usepackage{lineno}
\usepackage{comment}
\usepackage{array}  
\usepackage{siunitx}
\usepackage{mathtools}
\usepackage{calrsfs}
\usepackage{eucal}
\usepackage{bm}
\newcolumntype{d}{>{\displaystyle}c}

\usepackage{url}
\usepackage{hyperref}
\hypersetup{
	colorlinks=true,       
	linkcolor=red,          
	citecolor=green,        
	filecolor=magenta,      
	urlcolor=cyan           
}
\usepackage{enumitem}
\usepackage{pgfplots}

\usepackage{array}
\newcolumntype{L}[1]{>{\raggedright\let\newline\\\arraybackslash\hspace{0pt}}m{#1}}
\newcolumntype{C}[1]{>{\centering\let\newline\\\arraybackslash\hspace{0pt}}m{#1}}
\newcolumntype{R}[1]{>{\raggedleft\let\newline\\\arraybackslash\hspace{0pt}}m{#1}}

\title{Multi-target tracking for video surveillance using deep affinity network: A brief review}
\name{
	\begin{tabular}{ccc}
		 Sanam Nisar Mangi
	\end{tabular}
}

\address{
	\begin{tabular}{c}
		Norwegian University of Science and Technology, Norway.
	\end{tabular}
}

\begin{document}
%
\maketitle

\begin{abstract}
Deep learning models are known to function like the human brain. Due to their functional mechanism, they are frequently utilized to accomplish tasks that require human intelligence. Multi-target tracking (MTT) for video surveillance is one of the important and challenging tasks, which has attracted the researchers' attention due to its potential applications in various domains. Multi-target tracking tasks require locating the objects individually in each frame, which remains a huge challenge as there are immediate changes in appearances and extreme occlusions of objects. In addition to that, the Multi-target tracking framework requires multiple tasks to perform i.e. target detection, estimating trajectory, associations between frame, and re-identification. Various methods have been suggested, and some assumptions are made to constrain the problem in the context of a particular problem. In this paper, the state-of-the-art MTT models, which leverage from deep learning representational power are reviewed.
\end{abstract}

%


\section{Introduction:}

Multi-target tracking has found applications in diverse domains including surveillance systems \cite{Xiaogang,4}, anomaly detection \cite{66,67},  self-driving vehicles \cite{5,sultan1}, sports \cite{51,53}, robotics \cite{6}, segmentation \cite{68}\cite{69}, behaviour analysis \cite{Weiming}\cite{ullah2016}, virtual reality \cite{52}, and pose estimation \cite{3}\cite{54} etc. Although Multi-target tracking has been part of research studies for so many decades now, it's not fully optimized yet. The key issues that lead to complicated MTT tasks are frequent occlusions, path estimation, initialization and termination of a track, objects having similar appearances, and interaction between them. The multi-target tracking is divided into two main tasks i.e. object detection and tracking. MTT algorithms, to differentiate intra-group objects,  associate a unique id to each detected object that remains specific to the object for a certain time. These Ids are then utilized for generating motion trajectories for the tracked objects.  Detection is usually provided by a pre-trained detector and affinity model provides the estimation between the detections and an approach; optimizing association. The precision of the object detecting system determines how effective a target tracking system is.
The precision of the MTT models is highly influenced by factors like scale changes, frequent id switches, rotation, illumination variation, etc. Figure 1 illustrates the output of the MTT algorithm. Furthermore, there are the issues of background clutter, post shifting, track initialization and termination are complex tasks in multi-target tracking systems. To overcome these issues, researchers have taken advantage of deep neural networks, have proposed a variety of strategies.
\begin{figure*}
	\centering
	\includegraphics[scale=0.7]{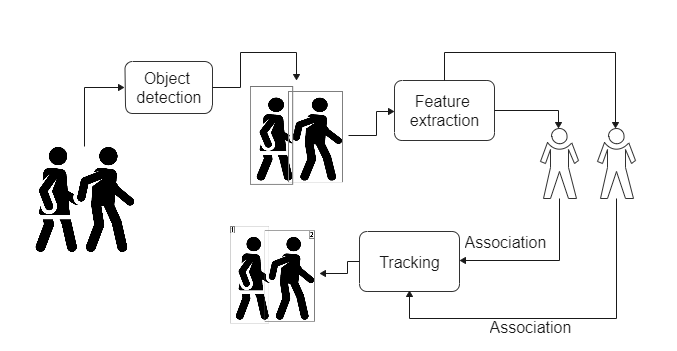}
	\caption{Illustrates the MTT algorithm's basic workflow} \label{fig1}
\end{figure*}

Since deep neural networks (DNNs) have the capability to retrieve abstract and complex features from input, by learning representations, they have been utilized in state-of-the-art methodologies to improve performance. In addition to extracting features, deep neural networks based architectures are designed for motion detection \cite{8} aiding to resolve the object affiliation task in tracking problems. This study will briefly review  state-of-the-art models that are proposed.

The organization of this analysis will be structured as,section 2 briefly review the previous work that has been done in terms of surveys or reviews, the MOT algorithms are briefly over viewed in section 3, followed by characterization of algorithms. Discussion of deep learning in multi-goal tracking is done in section 4. Section 5 covers the evaluation aspects metrics for assessment, datasets and discussion of benchmarks results, followed by conclusion.

\section{Related Work:}
In past studies, the problem of multi-target tracking has been formulated differently and from multiple perspectives. Several surveys and literature studies have been undertaken on those studies. Some of those reviews are dedicated to specific fields, such as crowd analysis \cite{56}\cite{9}\cite{54}, intelligent visual surveillance\cite{10}\cite{55},behaviour analysis\cite{11}\cite{12} etc. Some are dedicated to techniques e.g appearance models in visual tracking, Xi Li et al. in \cite{13}\cite{57} reviewed recent developments in 2D appearance models for visual object tracking. Alreshidi \cite{62} proposed hybrid features for facial emotion recognition but it could be used for multi object tracking as shown by Jiarui et al. \cite{63}. Nikolajs at el. \cite{18} identified the recent trends of Multi target tracking and determined the necessary computer vision building blocks for application in road traffic scenarios, including surveillance and planning. Ullah at el. \cite{57} evaluated different hand-crafted and learned features for the appearance model of the pedestrians. Wenhan Luo at el. \cite{14} investigated different important components involved in a MOT system as well as uncovered existing MOT research difficulties to present the first comprehensive review, deep learning-based strategies, on the other hand, were only briefly discussed because DNN is used by just a few algorithms. Patrick at el. \cite{15} reviewed learning based approaches however, they scoped their review around frameworks tackling data association in MTT tasks. Ullah at el. \cite{58}\cite{59} proposed a Bayesian framework and used hand-crafted and sparse coded deep features for modeling the appearance of the targets. The framework shows promising results when detectors gives accurate detection. Yingkun Xu at el. \cite{16} evaluated existing deep learning-based MOT approaches and discussed the performance improvement in deep learning methods. They also analyzed the mechanism of deep feature transferring, and rules to define a new tracking framework. Similarly, following the batch approach for multi-target tracking, Ullah and Faouzi \cite{60}\cite{61} proposed a deep learning based transportation network where the edge cost of the graph model through deep features and dynamic programming is used to optimize the trajectories of the individual objects. Gioele at el. \cite{17}  provided the first extensive study of Deep Learning's usage in Multiple Object Tracking, concentrating on 2D data retrieved from single-camera films. This survey also gathered experimental findings from the widely used MOT datasets and provided the numerical comparison between them in addition to making recommendations for possible future research paths.

\section{Multi-target tracking: An overview:}
The challenge of estimating the number of targets while maintaining their states or trajectories is defined as multi-target tracking (MTT). Various researchers have approached this topic in various ways over the years, and they have proposed algorithms accordingly. MTT models can be classified as following:

\subsection{Classification of MTT algorithms:}
Depending on how objects are initialized, MOT implementations can be classified as Detection based (DBT) or Detection free tracking (DFT). However, MTT models are standardized around Detection based training, where detections (identification of objects in a frame) are  retrieved as a pre-tracking step. Because an object-detector is required to identify the objects in DBT, performance is largely dependent on the detector's quality, hence choosing a detection framework is crucial. Some of the DFT and DBF algorithms are reviewed in the following section:

\subsubsection{Detection free tracking (DFT):} 
The output of the detector is often utilized as input to the tracker, and the tracker's output is provided to a motion forecasting algorithm that anticipates where objects will move in the following several seconds. However, that's not the case in detection free tracking. DFT based models require that a  fixed number of objects must be manually initialized in the first frame and these objects must then be localized in following frames. DFT is a difficult task since there is limited information about the object to be tracked, and that information is unclear. As a result, the initial bounding box merely approximates the object of interest from the background, and the object's appearance may change drastically over time \cite{21}. 

Luo at el. \cite{19} proposed a completely convolutional technique to develop a one-stage detector that uses spatio-temporal information acquired by a 3D sensor to conduct simultaneous 3D detection, tracking, and prediction. The model was said to function effectively in complex settings, tracking tiny and large moving vehicles accurately. Better feature representations were envisioned as a component of future effort. Katerina at el. \cite{20} proposed a method for video segmentation based on spectral clustering of trajectories, with the goal of segmenting and tracking closely interacting and defoaming agents in a video. This method first computes a video figure-ground segmentation, then assigns repulsive forces between foreground trajectories belonging to distinct connected components in any frame. Lu Zhang at el. \cite{21} developed a model-free tracker that tracks many objects at the same time by merging multiple single-object trackers with constraints on the objects' spatial organization. When spatial limitations between objects are taken into account, the performance improves. They also suggested that integrating structural limitations in DBT can help them perform better. Ming Yang at el. \cite{22} from a game-theoretic standpoint, gave a new viewpoint to MTT. They presented a decentralized MTT model in which a collection of independent kernel-based trackers are deployed, each of which contributes towards bridging the joint motion estimation and the Nash Equilibrium of a game. In both synthetic and actual video sequence, the proposed method achieved promising results. They include the idea of the incorporation of motion dynamic models in the trackers as a future work. Detection based tracking (DBT)
Object detection is used in DBT-based techniques to produce classification hypotheses for each object per frame, which are subsequently linked into trajectories. For these types of training models, the object detector is trained in advance, making the performance of such models highly dependent on the performance of the employed object detector. In addition to that, it makes DBT focus on the specific kind of targets such as pedestrians, vehicles, etc. Bastian Leibe at el. \cite{25}
suggested a model for dynamic scene interpretation from a moving vehicle that incorporates multi-view point and multi-category object detection for fully automatic scene geometry estimate, 2D object identification, 3D localization, trajectory estimation, and tracking. This study also proposed a novel tracking approach that uses a moving camera to locate and track a large number of objects. B. Bose at el. \cite{23} proposed a framework for the spatial proximity of objects, ideal tracks are often fragmented into components, or are merged with other tracks into common components. In this proposed method, the goal was to monitor a wide range of items in an open-world, unrestricted environment without requiring scene-specific training. This technique unifies fragments and merged track segments to build individual and full tracks for each object, employing foreground blobs obtained by background subtraction from stationary camera measurements. The tracking algorithm currently works in batch, but can be made online. Bi Song at el. \cite{24} studied the problem of long term tracking \& analyzes the statistical features and statistics of the targets along a proposed long-term track for each target, and proposes a stochastic graph evaluation framework to understand the link between tracklets in order to generate long term inter dependencies between the targets where the context information isn't readily available. Michael at el. \cite{26} proposed a unique approach for multi-person tracking by-detection in a particle filtering framework, employing an online trained classifier for data association, they investigated how unreliable information sources can be used for multi-target tracking.This method just employs 2D data; however, having more information about the scene, the target's appearance, and the camera motion would be advantageous. Luo at el. \cite{14} categorized  MOT algorithms as Online and offline tracking, based on the processing mode i.e. whether they rely on future frames when trying to decide the object identities in a given frame.

\subsubsection{Online tracking:}
Online tracking algorithms, also known as sequential tracking, generate predictions about the current frame based on information from the past and the present. This type of algorithm processes the frames in a step-wise manner. In some applications e.g. autonomous driving and robotics navigation, this information is prerequisite.

\subsubsection{Batch tracking:}
In order to determine the object identities in a given frame, batch tracking (offline tracking) techniques employ information from a previous frame. They frequently make use of global data, resulting in improved tracking quality; but, due to computational and memory constraints, it is not always possible to handle all of the frames at once.

\section{Deep Learning in Multi-target Tracking}
For multi-target tracking, deep learning algorithms have been shown to be effective, but there is still potential for improvement. Deep learning is used at various stages of the tracking process, therefore it is necessary to analyze and summarize how deep learning functionalities are contributing towards the performance improvement in each step.
Gioele at el. \cite{17} identified main steps that shared by majority of algorithms are as:

\subsection{Detection stage:}
Locating targets in a sequence of frames, using bounding boxes by analyzing input frames.

\subsection{Motion Prediction Stage:} Analyzing detections to extract appearance,motion or interaction features. 

\subsection{Affinity Stage:} Extracted features are utilized in similarity/distance computation between pair of detections.

\subsection{Association Stage:} The similarity/distance measures are utilized in  associating by providing the same ID to detection that correspond to the same target.

\subsection{Deep neural networks: Object detection Stage}
Object detection algorithms that are precise and fast are crucial because they allow computer vision applications to reach their full potential. Deep neural networks have been contributing in overall tracking performance which also includes specialized detection phases. Following are the examples of CNN-based models.

Redmon at el. \cite{27} proposed YOLO, single convolutional neural network simultaneously  predicts multiple bounding boxes and class probabilities directly from full images in one evaluation, trained on full images and directly optimizes detection performance, it also learns generalizable representations of objects. However, YOLO imposes strict spatial constraints on bounding box predictions, limiting the number of adjacent items that our model can predict. Small items that appear in groups, such as birds, are equally problematic for this model. Ren at el. \cite{28} proposed faster R-CNN, a single unified object identification network made up of fully deep CNN, improves detection accuracy and efficiency while reducing computational overhead. This model integrated a training method that alternates between fine-tuning for the region proposal and fine-tuning for the region proposal, enabling a unified, deep-learning-based object identification system to operate at frame rates that are close to real-time, then fine-tuning for object detection while maintaining the fixed targets. In certain surveillance settings, the occlusion is so frequent that it is not possible to detect the whole shape of the object like in the case of humans. To tackle this problem, Khan et al. \cite{64} proposed temporal consistency model that is trained to detect only the head locations. Similarly, techniques like \cite{65} is explored to do tracking of only the head locations rather than the whole body shape. Bewley at el {29} proposed framework  SORT to leverage the power of CNN based detection,  in the MOT prospect and achieve best in class performance with respect to both speed and accuracy,it focuses on frame-to-frame prediction and association.  It became capable of being rated as the best-performing by replacing the detection obtained from Aggregated Channel Features (ACF) with the detection computed by Faster R-CNN \cite{28}, based the architecture on Kalman filter \cite{30} and the Hungarian algorithm \cite{20}. CNN's has been used in the detection step for purposes other than constructing bounding boxes for objects in some cases. Min at el. \cite{31} suggested a new strategy,  for tracking multiple objects e.g. cars, with a combination of robust detection and two classifiers,For robust and precise identification of multiple vehicles, we propose the upgraded ViBe. It was used by CNN to eliminate false positives when the vibe Algorithm \cite{32} was used to recognize the car. It efficiently restrains dynamic noise and swiftly removes ghost shadows and object's leftover shadows.

\subsection{Deep neural networks: Feature Extraction and Motion Prediction }

Deep models have been utilized to extract meaningful features that are later used to associate detection outcomes, the features are learned through classification or recognition tasks, and they are inspired by the effectiveness of deep learning's representational capacity. Furthermore, it has been demonstrated e.g.Chu at el \cite{33}, Son at el \cite{34} that when MOT features such as temporal and spatial attention map or temporal order are investigated, performance can be improved. Additionally, some end-to-end deep learning based models are designed to extract the features not only for appearance descriptors but also for motion information.
Wang et al. \cite{33} proposed one of the earliest methods to apply DL in the MOT pipeline. This system takes advantage of the benefits of single object trackers to tackle the drifting difficulties caused by occlusion without compromising on the computational power.To enhance extracted characteristics, the network used encoders stacked in two layers, which were then used to compute affinity using SVM. The visibility map of the target is learned and then used to infer the spatial attention map, which is subsequently used to weight the features. In addition, the visibility map can be used to estimate occlusion status. This is what is known as a temporal attention process. The most commonly used CNN based approaches can be further classified as: Classical CNNs for features extraction and Siamese CNNs.
\subsubsection{Classical CNN for feature extraction:} 
Kim et al. \cite{36} claimed that the Multiple Hypotheses Tracking (MHT) technique is compatible with the existing visual tracking perspective. Modern advances in detection-based tracking and the development of efficient feature representations for object appearance have given the MHT process new possibilities. They improved MHT by integrating a regularized least squares framework for online training of appearance models for each track. Wojke et al. \cite{37} proposed to improve SORT\cite{29}, while it achieves favorable performance on high frame rates in terms of accuracy and precision, it yields a relatively large number of identity shifts. Wojke et al refine it by integrating appearance motion information, overcoming concerns by substituting the association metric with a convolutional neural network (CNN) that has been trained to distinguish pedestrians on a large scale person re-identification dataset. In comparison to sort, the updated tracking system effectively reduces the number of identity flips from 1423 to 781. This is a reduction of about 45 percent, allowing for competitive performance while maintaining real-time speed.

\subsubsection{Siamese CNNS:}
Siamese CNN has shown useful in MOT since the purpose of feature learning in the tracking phase is to determine the similarity between detection and tracks.
Leal-Taixe et al. \cite{38} implemented a two-stage strategy to match pairs of detection methods and brings a new perspective to object association challenges in the context of pedestrian tracking. They applied the CNN concept to multi-person tracking in this case and proposed to learn the decision whether two detections belong to the same trajectory,In order to avoid manual feature design for data association.The model's learning framework is divided into two stages. A CNN is pre-trained in Siamese twin architecture to measure the similarity of two equally sized image regions, and then the CNN is merged with the collected features to produce the prediction. They address the tracking problem well by formulating it as linear programming and combining deep features and motion information with a gradient boosting approach.

\subsection{Deep neural networks: Affinity computation stage:}
While some implementations use deep learning models to immediately produce an affinity score without the requirement for an explicit distance metric between features, other approaches calculate affinity between tracklets and detections by applying some distance measure to features obtained by a CNN.
Milan et al \cite{39} address the difficult problem of data association and trajectory estimation in a neural network environment. A recursive Bayesian filter, consisting of prediction and updating from observations, can be used to estimate the tracked object states in online MOT tasks, this model extends RNN to model this procedure, object states, existing observations, their corresponding matching matrix, as well as existence prospects are entered into the network as input. The predicted states and updated results are outputted, as well existence probabilities to decide whether the objects should be terminated, this model obtained promising tracking results. Chen et al \cite{40} proposed to compute affinity between the sampled particles and the tracked target, instead of being computed between targets and detections. Detections that did not coincide with the tracked objects were instead used to create new tracks and recover missing objects. Despite being an online monitoring algorithm, it was able to get top results on MOT15 at the time of publication, employing both public and private detections.

\subsection{Deep neural networks: Tracking/ Association step:}
Deep learning has been used in some MTT models to improve the association step. 
Ma et al.\cite{41} employed bi-directional GRU to decide where tracklets should be terminated as they expanded the Siamese tracklet network. For each detection, the network extracts tracklet features and sends them to a bidirectional GRU network.The output of the bidirectional GRU network is briefly pooled in Euclidean space to provide the overall features for tracklets. During the tracking process, tracklets are produced and then split into small subtracklets based on the local distance between bidirectional GRU outputs.Finally, considering similarities between global aspects of temporal pooling, these sub-tracklets are re-connected to long trajectories. On the MOT16 dataset, this approach obtains results that are equivalent to the state of the art.
Ren et al. \cite{42} propose a collaborative implementation in which many deep RL agents were used to accomplish the association task. A prediction network and a decision network were the two key components of the model. Using recent tracklet trajectory, the CNN was employed as a prediction network and was trained to anticipate target movement in new frames.

\subsection{Deep neural networks: Other uses in MTT}
Few of the models are briefly defined in this section, that do not fit in those four identified steps.
Jiang et al \cite{43} used a Deep RL agent to accomplish bounding box regression, to the employed tracking algorithms, to increase efficiency.VGG-16 CNN was employed to to extract appearance, those extracted features were stored to keep and use the history of last 10 movements taken by the target, and then the integrated network predicts one of several alternative outcomes, including bounding box motion and scaling, as well as termination action. On the MOT15 dataset, using this bounding box regression methodology on several state-of-the-art MOT algorithms resulted in improvements of 2 to 7 absolute MOTA points, placing it first among public detection approaches.
Xiang et al.\cite{44} deployed MetricNet for pedestrian tracking, combining an affinity model with trajectory estimation obtained by  Bayesian filters. The features were extracted using a VGG-16 CNN trained for target re-identification, extracted features and performed bounding box regression, the motion model has two parts, one takes trajectory coordinates as input, the other combine the extracted features and a detection box that performs the Bayesian filtering and outputs the target's updated position On MOT16 and MOT15, the algorithm delivered the best and second-best scores among online approaches, respectively.
Recent advances in model free single object tracking (SOT) algorithms have significantly pushed the use of SOT in multi-object tracking (MOT) to improve resilience and reduce dependency on external detectors. SOT algorithms, on the other hand, are typically designed to distinguish a target from its surroundings, and they frequently encounter issues when a target is spatially mixed with similar artifacts, as seen in MOT.
Chu et al. \cite{45} proposed a model to address the issue of robustness and eliminating dependency over the external detectors. They implemented a model using three different CNNs in their algorithm. The  PafNet is integrated to differentiate between the background and the tracked objects. The PartNet differentiate between tracked targets.The other integrated CNN is the convolutional layer, which determines whether or not the tracking model should be refreshed. Using an SVM classifier and the Hungarian technique, unassociated detections were used to recover from target occlusion. The algorithm was tested on the MOT15 and MOT16 datasets, with the first yielding the best overall results and the second yielding the best results among online approaches.

\section{MTT Evaluation}
Target tracking has been an aspect of research for many years, and all of the proposed methods are subjected to a quantitative evaluation to determine their efficacy, which is necessary to compute the impacts of various components and to compare them to other approaches. There are following components for MTT evaluations.

\subsection{Metrics:}
Set of standard metrics are available to facilitate comparisons of algorithms in experimental setup, to evaluate the performance and indicate the drawbacks. The most relevant are Classical metrics and CLEAR MOT metrics \cite{46}\cite{47}. Classical metrics point out faults that the algorithm can come across for-example, Mostly Tracked (MT) trajectories,Mostly Lost (ML) trajectories, ID switches etc. CLEAR MOT metrics are MOTA (Multiple Object Tracking Accuracy ) and MOTP (Multiple Object Tracking Precision). MOTA combines the false positive, false negative, and mismatch rates into a single value, resulting in a generally good overall tracking performance. Despite several flaws and complaints, this is by far the most extensively used assessment method. MOTP describes how precisely objects are tracked using bounding box overlap and/or distance measurements.

\begin{table}
	\centering
	\caption{The average fragmentation score produced by an online and offline as given by Gioele at el. \cite{17}.}\label{tab1}
	\begin{tabular}{|l|l|l|l|}
		\hline
		Mode &  MOT17 & MOT16 & MOT17\\
		\hline
		Batch &  {\bfseries 1143.8} & {\bfseries 1104.9}  & {\bfseries 3188.2}\\
		Online &  {\bfseries 1509.5} & {\bfseries 1820.2}  & {\bfseries 7555.8}\\
		\hline
	\end{tabular}
\end{table}

\subsection{Benchmark Datasets:}
Benchmark Datasets include MOTChallenge, KITTI, UA-DETRAC. The MOTChallenge datasets are the largest and most complete datasets for pedestrian tracking currently available, offering more data to train deep models. MOT15 \cite{48} is the initial MOT Challenge dataset, and it features video with a range of attributes that the model would need to generalize better in order to achieve good results. MOT16 \cite{7} and MOT19 \cite{35} are the other modified versions.

\subsection{Benchmark Results:}
Gioele et al. \cite{17} list the public results tested on MOT Challenge MOT15 datasets and MOT16 data-sets, documented from the corresponding publication, to achieve a clear comparison of results between the approaches mentioned in this work.
The findings are classified into public and private detection-based models since the quality of detection has an impact on performance. The methods are divided into two categories: online and offline.
The year of the published referenced document, its operating mode, the MOTA, MOTP, IDF1, Mostly Tracked (MT) and Mostly Lost (ML) metrics, expressed in percentages; the absolute number of false positives (FP), false negatives (FN), ID switches (IDS), and fragmentations (Frag); the algorithm's speed expressed in frames per second (Hz). For each measure, an arrow pointing up (↑) implies a higher score, while an arrow pointing down (↓) suggests the reverse. The best performance is emphasized among models that operate in the same mode (batch/online), and each statistic is highlighted in bold. We only listed the results obtained from models that were visited in this review in table. \ref{tab2} and \ref{tab3}.

\begin{table*}
	\caption{Experimental results \cite{17} of deep learning MOT algorithms on MOT15  }
	\label{tab2}
	\begin{tabular}{||c c c c c c c c c c c c c||} 
		\hline
		& Year & Mode & MOTA $\uparrow$ & MOTP$\uparrow$ & IDF1$\uparrow$ & MT$\uparrow$ & ML$\downarrow$ & FP$\downarrow$ & FN$\downarrow$ & IDS$\downarrow$ & Frag$\downarrow$ & Hz$\uparrow$ \\ [0.5ex] 
		\hline\hline
		\cite{36} & 2015 & Online & 32.4 & 71.8 & 45.3 & 16.0 & 43.8 & 9064 &  32060 & \textbf{435} & \underline{826} & 0.7 \\ 
		
		\hline
		\cite{39} & 2017 & Online & 19.0 & 71.0 & 17.1 & 5.5 & 45.6 & 11578 &  36706 & 1490 & 2081 & \textbf{165.2} \\
		\hline
		
		\cite{40} & 2017 & Online & 38.5 & \underline{72.6} & 47.1 & 8.7 & 37.4 & \textbf{4005} &  33204 & 586 & 1263 & 6.7 \\
		\hline
		
		\hline
		\cite{44} & 2019 & Online & 37.1 & 72.5 & \textbf{48.4} & 12.6 & 39.7 & 8305 & 29732 & 580 & 1193 & 1.0 \\
		\hline
		\cite{33} & 2019 & Online & 38.9 & 70.6 & 44.7 & \textbf{16.6} & 31.5 & 7321 & 29501  & 720 & 1440 & 0.3 \\
		\hline
		\hline
		\cite{49} & 2019 & Batch & 28.1 & \textbf{74.3} & 38.7 &  &  & 6733 & 36952  & \underline{477} & \textbf{790} & 16.9 \\
		
		\hline
	\end{tabular}
\end{table*}

\begin{table*}
	\caption{Experimental results \cite{17} of deep learning MOT algorithms on MOT16 }
	\label{tab3}
	\begin{tabular}{||c c c c c c c c c c c c c||} 
		\hline
		& Year & Mode & MOTA $\uparrow$ & MOTP$\uparrow$ & IDF1$\uparrow$ & MT$\uparrow$ & ML$\downarrow$ & FP$\downarrow$ & FN$\downarrow$ & IDS$\downarrow$ & Frag$\downarrow$ & Hz$\uparrow$ \\ [0.5ex] 
		\hline\hline
		
		\cite{3} & 2017 & Online & 46.0 & 74.9 & 50.0 & 14.6 & 43.6 & 6895 &  91117 & \underline{473} & 1422 & \textbf{0.2} \\
		\hline
		
		\cite{42} & 2018 & Online & 47.3 & 74.6 &  & \underline{17.4} & 39.9 & 6375 &  88543 &  &  &  \\
		\hline
		
		\hline
		\cite{45} & 2019 & Online & \underline{48.8} & 75.7 & 47.2 & 15.8 & 38.1 & 5875 & 86567 & 906 & 1116 & 1.0 \\
		\hline
		\cite{45} & 2019 & Online &  \underline{48.8} & 75.7 & 47.2 & 15.8 &  \underline{38.1} & 5875 &  86567 & 906 & 1116 & 0.1 \\ 
		
		\hline
	\end{tabular}
\end{table*}

In reality, the models that use deep learning and have online processing mode produced the greatest results.This could, however, be the result of a stronger emphasis on building online approaches, which is becoming more popular in the MOT deep learning research community. The larger number of fragmentations is a frequent problem among online approaches that is not reflected in the MOTA score. When occlusions or detections are missing, online algorithms do not look ahead, re-identify the missing targets, or interpolate the missing piece of the trajectories in the video.

\section{Conclusion:}
The methods that use deep learning to tackle the MTT problem have been briefly explored in this study. This study has discussed solutions that use deep learning to address each of the four steps of the MTT issue, bringing the total number of state-of-the-art MOT techniques to n. MOT algorithm evaluations, including evaluation measures and benchmark results from accessible data sets, were briefly discussed. Single object trackers have recently benefited from the introduction of deep models into global graph optimization algorithms, resulting in high-performing online trackers; batch techniques, on the other hand, have benefited from the introduction of deep models into global graph optimization algorithms.



{\small
\bibliographystyle{IEEEbib}
\bibliography{tt}

\begin{thebibliography}{10}

\bibitem{Xiaogang}
Xiaogang Wang,
\newblock ``Intelligent multi-camera video surveillance: A review,''
\newblock {\em Pattern recognition letters}, vol. 34, no. 1, pp. 3--19, 2013.

\bibitem{4}
Samuel Blackman and Robert Popoli,
\newblock ``Design and analysis of modern tracking systems(book),''
\newblock {\em Norwood, MA: Artech House, 1999.}, 1999.

\bibitem{66}
Habib Ullah, Muhammad Uzair, Mohib Ullah, Asif Khan, Ayaz Ahmad, and Wilayat
  Khan,
\newblock ``Density independent hydrodynamics model for crowd coherency
  detection,''
\newblock {\em Neurocomputing}, vol. 242, pp. 28--39, 2017.

\bibitem{67}
Habib Ullah, Ahmed~B Altamimi, Muhammad Uzair, and Mohib Ullah,
\newblock ``Anomalous entities detection and localization in pedestrian
  flows,''
\newblock {\em Neurocomputing}, vol. 290, pp. 74--86, 2018.

\bibitem{5}
Dequan Zeng, Lu~Xiong, Zhuoping Yu, Qiping Chen, Zhiqiang Fu, Zhuoren Li,
  Peizhi Zhang, Puhang Xu, Zixuan Qian, Hongyu Xiao, et~al.,
\newblock ``A priority data association policy for multitarget tracking on
  intelligent vehicle risk assessment,''
\newblock {\em Remote Sensing}, vol. 12, no. 19, pp. 3255, 2020.

\bibitem{sultan1}
Sultan~Daud Khan and Habib Ullah,
\newblock ``A survey of advances in vision-based vehicle re-identification,''
\newblock {\em Computer Vision and Image Understanding}, vol. 182, pp. 50--63,
  2019.

\bibitem{51}
Sultan~Daud Khan, Habib Ullah, Mohib Ullah, Nicola Conci, Faouzi~Alaya Cheikh,
  and Azeddine Beghdadi,
\newblock ``Person head detection based deep model for people counting in
  sports videos,''
\newblock in {\em 2019 16th IEEE International Conference on Advanced Video and
  Signal Based Surveillance (AVSS)}. IEEE, 2019, pp. 1--8.

\bibitem{53}
Mohib Ullah, Muhammad Mudassar~Yamin, Ahmed Mohammed, Sultan Daud~Khan, Habib
  Ullah, and Faouzi Alaya~Cheikh,
\newblock ``Attention-based lstm network for action recognition in sports,''
\newblock {\em Electronic Imaging}, vol. 2021, no. 6, pp. 302--1, 2021.

\bibitem{6}
Jacopo Banfi, J{\'e}r{\^o}me Guzzi, Alessandro Giusti, Luca Gambardella, and
  Gianni~A Di~Caro,
\newblock ``Fair multi-target tracking in cooperative multi-robot systems,''
\newblock in {\em 2015 IEEE International Conference on Robotics and Automation
  (ICRA)}. IEEE, 2015, pp. 5411--5418.

\bibitem{68}
Habib Ullah, Mohib Ullah, and Muhammad Uzair,
\newblock ``A hybrid social influence model for pedestrian motion
  segmentation,''
\newblock {\em Neural Computing and Applications}, vol. 31, no. 11, pp.
  7317--7333, 2019.

\bibitem{69}
Zhenbo Xu, Wei Zhang, Xiao Tan, Wei Yang, Huan Huang, Shilei Wen, Errui Ding,
  and Liusheng Huang,
\newblock ``Segment as points for efficient online multi-object tracking and
  segmentation,''
\newblock in {\em European Conference on Computer Vision}. Springer, 2020, pp.
  264--281.

\bibitem{Weiming}
Weiming Hu, Tieniu Tan, Liang Wang, and Steve Maybank,
\newblock ``A survey on visual surveillance of object motion and behaviors,''
\newblock {\em IEEE Transactions on Systems, Man, and Cybernetics, Part C
  (Applications and Reviews)}, vol. 34, no. 3, pp. 334--352, 2004.

\bibitem{ullah2016}
Mohib Ullah, Habib Ullah, Nicola Conci, and Francesco~GB De~Natale,
\newblock ``Crowd behavior identification,''
\newblock in {\em 2016 IEEE international conference on image processing
  (ICIP)}. IEEE, 2016, pp. 1195--1199.

\bibitem{52}
Habib Ullah, Sultan~Daud Khan, Mohib Ullah, and Faouzi~Alaya Cheikh,
\newblock ``Social modeling meets virtual reality: An immersive implication,''
\newblock in {\em International Conference on Pattern Recognition}. Springer,
  2021, pp. 131--140.

\bibitem{3}
Peng Chu, Heng Fan, Chiu~C Tan, and Haibin Ling,
\newblock ``Online multi-object tracking with instance-aware tracker and
  dynamic model refreshment,''
\newblock in {\em 2019 IEEE Winter Conference on Applications of Computer
  Vision (WACV)}. IEEE, 2019, pp. 161--170.

\bibitem{54}
Akif~Quddus Khan, Salman Khan, Mohib Ullah, and Faouzi~Alaya Cheikh,
\newblock ``A bottom-up approach for pig skeleton extraction using rgb data,''
\newblock in {\em International Conference on Image and Signal Processing}.
  Springer, 2020, pp. 54--61.

\bibitem{8}
Yingkun Xu, Xiaolong Zhou, Shengyong Chen, and Fenfen Li,
\newblock ``Deep learning for multiple object tracking: a survey,''
\newblock {\em IET Computer Vision}, vol. 13, no. 4, pp. 355--368, 2019.

\bibitem{56}
Sultan~Daud Khan, Maqsood Mahmud, Habib Ullah, Mohib Ullah, and Faouzi~Alaya
  Cheikh,
\newblock ``Crowd congestion detection in videos,''
\newblock {\em Electronic Imaging}, vol. 2020, no. 6, pp. 72--1, 2020.

\bibitem{9}
Beibei Zhan, Dorothy~N Monekosso, Paolo Remagnino, Sergio~A Velastin, and
  Li-Qun Xu,
\newblock ``Crowd analysis: a survey,''
\newblock {\em Machine Vision and Applications}, vol. 19, no. 5, pp. 345--357,
  2008.

\bibitem{10}
In~Su Kim, Hong~Seok Choi, Kwang~Moo Yi, Jin~Young Choi, and Seong~G Kong,
\newblock ``Intelligent visual surveillance—a survey,''
\newblock {\em International Journal of Control, Automation and Systems}, vol.
  8, no. 5, pp. 926--939, 2010.

\bibitem{55}
Habib Ullah, Mohib Ullah, and Muhammad Uzair,
\newblock ``A hybrid social influence model for pedestrian motion
  segmentation,''
\newblock {\em Neural Computing and Applications}, vol. 31, no. 11, pp.
  7317--7333, 2019.

\bibitem{11}
Joshua Candamo, Matthew Shreve, Dmitry~B Goldgof, Deborah~B Sapper, and
  Rangachar Kasturi,
\newblock ``Understanding transit scenes: A survey on human
  behavior-recognition algorithms,''
\newblock {\em IEEE transactions on intelligent transportation systems}, vol.
  11, no. 1, pp. 206--224, 2009.

\bibitem{12}
David~A Forsyth, Okan Arikan, Leslie Ikemoto, Deva Ramanan, and James O'Brien,
\newblock ``Computational studies of human motion: Tracking and motion
  synthesis,''
\newblock 2006.

\bibitem{13}
Xi~Li, Weiming Hu, Chunhua Shen, Zhongfei Zhang, Anthony Dick, and Anton
  Van~Den Hengel,
\newblock ``A survey of appearance models in visual object tracking,''
\newblock {\em ACM transactions on Intelligent Systems and Technology (TIST)},
  vol. 4, no. 4, pp. 1--48, 2013.

\bibitem{57}
Mohib Ullah, Habib Ullah, and Faouzi~Alaya Cheikh,
\newblock ``Single shot appearance model (ssam) for multi-target tracking,''
\newblock {\em Electronic Imaging}, vol. 2019, no. 7, pp. 466--1, 2019.

\bibitem{62}
Abdulrahman Alreshidi and Mohib Ullah,
\newblock ``Facial emotion recognition using hybrid features,''
\newblock in {\em Informatics}. Multidisciplinary Digital Publishing Institute,
  2020, vol.~7, p.~6.

\bibitem{63}
Jiarui Xu, Yue Cao, Zheng Zhang, and Han Hu,
\newblock ``Spatial-temporal relation networks for multi-object tracking,''
\newblock in {\em Proceedings of the IEEE/CVF International Conference on
  Computer Vision}, 2019, pp. 3988--3998.

\bibitem{18}
Nikolajs Bumanis, Gatis Vitols, Irina Arhipova, and Egons Solmanis,
\newblock ``Multi-object tracking for urban and multilane traffic: Building
  blocks for real-world application,''
\newblock 2021.

\bibitem{14}
Wenhan Luo, Junliang Xing, Anton Milan, Xiaoqin Zhang, Wei Liu, and Tae-Kyun
  Kim,
\newblock ``Multiple object tracking: A literature review,''
\newblock {\em Artificial Intelligence}, p. 103448, 2020.

\bibitem{15}
Patrick Emami, Panos~M Pardalos, Lily Elefteriadou, and Sanjay Ranka,
\newblock ``Machine learning methods for data association in multi-object
  tracking,''
\newblock {\em ACM Computing Surveys (CSUR)}, vol. 53, no. 4, pp. 1--34, 2020.

\bibitem{58}
Mohib Ullah, Mohammed~Ahmed Kedir, and Faouzi~Alaya Cheikh,
\newblock ``Hand-crafted vs deep features: A quantitative study of pedestrian
  appearance model,''
\newblock in {\em 2018 Colour and Visual Computing Symposium (CVCS)}. IEEE,
  2018, pp. 1--6.

\bibitem{59}
Mohib Ullah, Faouzi~Alaya Cheikh, and Ali~Shariq Imran,
\newblock ``Hog based real-time multi-target tracking in bayesian framework,''
\newblock in {\em 2016 13th IEEE International Conference on Advanced Video and
  Signal Based Surveillance (AVSS)}. IEEE, 2016, pp. 416--422.

\bibitem{16}
Yingkun Xu, Xiaolong Zhou, Shengyong Chen, and Fenfen Li,
\newblock ``Deep learning for multiple object tracking: a survey,''
\newblock {\em IET Computer Vision}, vol. 13, no. 4, pp. 355--368, 2019.

\bibitem{60}
Mohib Ullah and Faouzi~Alaya Cheikh,
\newblock ``Deep feature based end-to-end transportation network for
  multi-target tracking,''
\newblock in {\em 2018 25th IEEE International Conference on Image Processing
  (ICIP)}. IEEE, 2018, pp. 3738--3742.

\bibitem{61}
Mohib Ullah and Faouzi Alaya~Cheikh,
\newblock ``A directed sparse graphical model for multi-target tracking,''
\newblock in {\em Proceedings of the IEEE Conference on Computer Vision and
  Pattern Recognition Workshops}, 2018, pp. 1816--1823.

\bibitem{17}
Gioele Ciaparrone, Francisco~Luque S{\'a}nchez, Siham Tabik, Luigi Troiano,
  Roberto Tagliaferri, and Francisco Herrera,
\newblock ``Deep learning in video multi-object tracking: A survey,''
\newblock {\em Neurocomputing}, vol. 381, pp. 61--88, 2020.

\bibitem{21}
Lu~Zhang and Laurens van~der Maaten,
\newblock ``Structure preserving object tracking,''
\newblock in {\em Proceedings of the IEEE conference on computer vision and
  pattern recognition}, 2013, pp. 1838--1845.

\bibitem{19}
Wenjie Luo, Bin Yang, and Raquel Urtasun,
\newblock ``Fast and furious: Real time end-to-end 3d detection, tracking and
  motion forecasting with a single convolutional net,''
\newblock in {\em Proceedings of the IEEE conference on Computer Vision and
  Pattern Recognition}, 2018, pp. 3569--3577.

\bibitem{20}
Katerina Fragkiadaki and Jianbo Shi,
\newblock ``Detection free tracking: Exploiting motion and topology for
  segmenting and tracking under entanglement,''
\newblock in {\em CVPR 2011}. IEEE, 2011, pp. 2073--2080.

\bibitem{22}
Ming Yang, Ting Yu, and Ying Wu,
\newblock ``Game-theoretic multiple target tracking,''
\newblock in {\em 2007 IEEE 11th International Conference on Computer Vision}.
  IEEE, 2007, pp. 1--8.

\bibitem{25}
Bastian Leibe, Nico Cornelis, Kurt Cornelis, and Luc Van~Gool,
\newblock ``Dynamic 3d scene analysis from a moving vehicle,''
\newblock in {\em 2007 IEEE Conference on Computer Vision and Pattern
  Recognition}. IEEE, 2007, pp. 1--8.

\bibitem{23}
Biswajit Bose, Xiaogang Wang, and Eric Grimson,
\newblock ``Multi-class object tracking algorithm that handles fragmentation
  and grouping,''
\newblock in {\em 2007 IEEE Conference on Computer Vision and Pattern
  Recognition}. IEEE, 2007, pp. 1--8.

\bibitem{24}
Bi~Song, Ting-Yueh Jeng, Elliot Staudt, and Amit~K Roy-Chowdhury,
\newblock ``A stochastic graph evolution framework for robust multi-target
  tracking,''
\newblock in {\em European Conference on Computer Vision}. Springer, 2010, pp.
  605--619.

\bibitem{26}
Michael~D Breitenstein, Fabian Reichlin, Bastian Leibe, Esther Koller-Meier,
  and Luc Van~Gool,
\newblock ``Robust tracking-by-detection using a detector confidence particle
  filter,''
\newblock in {\em 2009 IEEE 12th International Conference on Computer Vision}.
  IEEE, 2009, pp. 1515--1522.

\bibitem{27}
Joseph Redmon, Santosh Divvala, Ross Girshick, and Ali Farhadi,
\newblock ``You only look once: Unified, real-time object detection,''
\newblock in {\em Proceedings of the IEEE conference on computer vision and
  pattern recognition}, 2016, pp. 779--788.

\bibitem{28}
Shaoqing Ren, Kaiming He, Ross Girshick, and Jian Sun,
\newblock ``Faster r-cnn: Towards real-time object detection with region
  proposal networks,''
\newblock {\em Advances in neural information processing systems}, vol. 28, pp.
  91--99, 2015.

\bibitem{64}
Sultan~Daud Khan, Ahmed~B Altamimi, Mohib Ullah, Habib Ullah, and Faouzi~Alaya
  Cheikh,
\newblock ``Tcm: Temporal consistency model for head detection in complex
  videos,''
\newblock {\em Journal of Sensors}, vol. 2020, 2020.

\bibitem{65}
Mohib Ullah, Maqsood Mahmud, Habib Ullah, Kashif Ahmad, Ali~Shariq Imran, and
  Faouzi~Alaya Cheikh,
\newblock ``Head based tracking,''
\newblock {\em Electronic Imaging}, vol. 2020, no. 6, pp. 74--1, 2020.

\bibitem{30}
Rudolph~Emil Kalman,
\newblock ``A new approach to linear filtering and prediction problems,''
\newblock 1960.

\bibitem{31}
Weidong Min, Mengdan Fan, Xiaoguang Guo, and Qing Han,
\newblock ``A new approach to track multiple vehicles with the combination of
  robust detection and two classifiers,''
\newblock {\em IEEE Transactions on Intelligent Transportation Systems}, vol.
  19, no. 1, pp. 174--186, 2017.

\bibitem{32}
Olivier Barnich and Marc Van~Droogenbroeck,
\newblock ``Vibe: A universal background subtraction algorithm for video
  sequences,''
\newblock {\em IEEE Transactions on Image processing}, vol. 20, no. 6, pp.
  1709--1724, 2010.

\bibitem{33}
Qi~Chu, Wanli Ouyang, Hongsheng Li, Xiaogang Wang, Bin Liu, and Nenghai Yu,
\newblock ``Online multi-object tracking using cnn-based single object tracker
  with spatial-temporal attention mechanism,''
\newblock in {\em Proceedings of the IEEE international conference on computer
  vision}, 2017, pp. 4836--4845.

\bibitem{34}
Jeany Son, Mooyeol Baek, Minsu Cho, and Bohyung Han,
\newblock ``Multi-object tracking with quadruplet convolutional neural
  networks,''
\newblock in {\em Proceedings of the IEEE conference on computer vision and
  pattern recognition}, 2017, pp. 5620--5629.

\bibitem{36}
Chanho Kim, Fuxin Li, Arridhana Ciptadi, and James~M Rehg,
\newblock ``Multiple hypothesis tracking revisited,''
\newblock in {\em Proceedings of the IEEE international conference on computer
  vision}, 2015, pp. 4696--4704.

\bibitem{37}
Alex Bewley, Zongyuan Ge, Lionel Ott, Fabio Ramos, and Ben Upcroft,
\newblock ``Simple online and realtime tracking,''
\newblock in {\em 2016 IEEE international conference on image processing
  (ICIP)}. IEEE, 2016, pp. 3464--3468.

\bibitem{29}
Alex Bewley, Zongyuan Ge, Lionel Ott, Fabio Ramos, and Ben Upcroft,
\newblock ``Simple online and realtime tracking,''
\newblock in {\em 2016 IEEE international conference on image processing
  (ICIP)}. IEEE, 2016, pp. 3464--3468.

\bibitem{38}
Laura Leal-Taix{\'e}, Cristian Canton-Ferrer, and Konrad Schindler,
\newblock ``Learning by tracking: Siamese cnn for robust target association,''
\newblock in {\em Proceedings of the IEEE Conference on Computer Vision and
  Pattern Recognition Workshops}, 2016, pp. 33--40.

\bibitem{39}
Anton Milan, S~Hamid Rezatofighi, Anthony Dick, Ian Reid, and Konrad Schindler,
\newblock ``Online multi-target tracking using recurrent neural networks,''
\newblock in {\em Thirty-First AAAI Conference on Artificial Intelligence},
  2017.

\bibitem{40}
Long Chen, Haizhou Ai, Chong Shang, Zijie Zhuang, and Bo~Bai,
\newblock ``Online multi-object tracking with convolutional neural networks,''
\newblock in {\em 2017 IEEE International Conference on Image Processing
  (ICIP)}. IEEE, 2017, pp. 645--649.

\bibitem{41}
Cong Ma, Changshui Yang, Fan Yang, Yueqing Zhuang, Ziwei Zhang, Huizhu Jia, and
  Xiaodong Xie,
\newblock ``Trajectory factory: Tracklet cleaving and re-connection by deep
  siamese bi-gru for multiple object tracking,''
\newblock in {\em 2018 IEEE International Conference on Multimedia and Expo
  (ICME)}. IEEE, 2018, pp. 1--6.

\bibitem{42}
Liangliang Ren, Jiwen Lu, Zifeng Wang, Qi~Tian, and Jie Zhou,
\newblock ``Collaborative deep reinforcement learning for multi-object
  tracking,''
\newblock in {\em Proceedings of the European Conference on Computer Vision
  (ECCV)}, 2018, pp. 586--602.

\bibitem{43}
Yifan Jiang, Hyunhak Shin, and Hanseok Ko,
\newblock ``Precise regression for bounding box correction for improved
  tracking based on deep reinforcement learning,''
\newblock in {\em 2018 IEEE International Conference on Acoustics, Speech and
  Signal Processing (ICASSP)}. IEEE, 2018, pp. 1643--1647.

\bibitem{44}
Liqian Ma, Siyu Tang, Michael~J Black, and Luc Van~Gool,
\newblock ``Customized multi-person tracker,''
\newblock in {\em Asian conference on computer vision}. Springer, 2018, pp.
  612--628.

\bibitem{45}
Peng Chu, Heng Fan, Chiu~C Tan, and Haibin Ling,
\newblock ``Online multi-object tracking with instance-aware tracker and
  dynamic model refreshment,''
\newblock in {\em 2019 IEEE Winter Conference on Applications of Computer
  Vision (WACV)}. IEEE, 2019, pp. 161--170.

\bibitem{46}
Keni Bernardin and Rainer Stiefelhagen,
\newblock ``Evaluating multiple object tracking performance: the clear mot
  metrics,''
\newblock {\em EURASIP Journal on Image and Video Processing}, vol. 2008, pp.
  1--10, 2008.

\bibitem{47}
Branko Ristic, Ba-Ngu Vo, Daniel Clark, and Ba-Tuong Vo,
\newblock ``A metric for performance evaluation of multi-target tracking
  algorithms,''
\newblock {\em IEEE Transactions on Signal Processing}, vol. 59, no. 7, pp.
  3452--3457, 2011.

\bibitem{48}
Laura Leal-Taix{\'e}, Anton Milan, Ian Reid, Stefan Roth, and Konrad Schindler,
\newblock ``Motchallenge 2015: Towards a benchmark for multi-target tracking,''
\newblock {\em arXiv preprint arXiv:1504.01942}, 2015.

\bibitem{7}
Andrii Maksai and Pascal Fua,
\newblock ``Eliminating exposure bias and loss-evaluation mismatch in multiple
  object tracking,''
\newblock {\em arXiv preprint arXiv:1811.10984}, 2018.

\bibitem{35}
Patrick Dendorfer, Hamid Rezatofighi, Anton Milan, Javen Shi, Daniel Cremers,
  Ian Reid, Stefan Roth, Konrad Schindler, and Laura Leal-Taixe,
\newblock ``Cvpr19 tracking and detection challenge: How crowded can it get?,''
\newblock {\em arXiv preprint arXiv:1906.04567}, 2019.

\bibitem{49}
Long Chen, Haizhou Ai, Rui Chen, and Zijie Zhuang,
\newblock ``Aggregate tracklet appearance features for multi-object tracking,''
\newblock {\em IEEE Signal Processing Letters}, vol. 26, no. 11, pp.
  1613--1617, 2019.

\end{thebibliography}
}
\end{document}